\documentclass[10pt,twocolumn,letterpaper]{article}

\usepackage{iccv}
\usepackage{times}
\usepackage{epsfig}
\usepackage{graphicx}
\usepackage{amsmath}
\usepackage{amssymb}
\usepackage{mathtools}
\usepackage{multirow}
\usepackage{makecell}
\usepackage{booktabs}
\usepackage{tabularx}
\usepackage[super]{nth}
\usepackage{paralist}
\usepackage{algorithm2e}
\usepackage{balance}
\usepackage{multibib}
\usepackage{titling}

\makeatletter
\newcommand{\removelatexerror}{\let\@latex@error\@gobble}
\makeatother

\usepackage{helvet} 
\usepackage{floatrow}
\newfloatcommand{capbtabbox}{table}[][\FBwidth]

\usepackage[format=plain,labelformat=simple,labelsep=period,font=small,skip=4pt,compatibility=false]{caption}
\usepackage[font=scriptsize,skip=2pt,subrefformat=parens]{subcaption}

\newcommand{\norm}[1]{\left\lVert#1\right\rVert}
\newcommand*{\myparagraph}[1]{\vspace{0.5em}\noindent\textbf{#1}}

\DeclareMathOperator*{\argmax}{arg\,max}

\usepackage{xspace}

\newcommand{\Eq}{Eq.\@\xspace}

\newcommand{\tssim}{\textit{t-}DSSIM\xspace}
\newcommand{\tpsnr}{\textit{t-}PSNR\xspace}

\usepackage{color}
\definecolor{mygreen}{rgb}{0.2, 0.7, 0.1}
\usepackage[pagebackref=true,breaklinks=true,colorlinks,citecolor={mygreen},bookmarks=false]{hyperref}

\pretitle{\begin{center}\Large \bf}
\posttitle{\par \end{center}}

\usepackage[capitalize]{cleveref}
\crefname{section}{Sec.}{Section}

\iccvfinalcopy 

\ificcvfinal\fi
\newcites{supp}{References}

\usepackage{fancyhdr}
\usepackage{setspace}

\begin{document}

\title{Markov Decision Process for Video Generation}
\author{Vladyslav Yushchenko$^1$\thanks{This work was done while VY was at TU Darmstadt.} \qquad Nikita Araslanov$^2$ \qquad Stefan Roth$^2$\\
$^1$iNTENCE automotive electronics GmbH, $^2$TU Darmstadt
}

\setlength{\thanksmarkwidth}{1.45em}

\date{}

\maketitle

\pagestyle{empty}
\thispagestyle{fancy}

\begin{abstract}
We identify two pathological cases of temporal inconsistencies in video generation: video freezing and video looping.
To better quantify the temporal diversity, we propose a class of complementary metrics that are effective, easy to implement, data agnostic, and interpretable.
Further, we observe that current state-of-the-art models are trained on video samples of fixed length thereby inhibiting long-term modeling.
To address this, we reformulate the problem of video generation as a Markov Decision Process (MDP).
The underlying idea is to represent motion as a stochastic process with an infinite forecast horizon to overcome the fixed length limitation and to mitigate the presence of temporal artifacts.
We show that our formulation is easy to integrate into the state-of-the-art MoCoGAN framework.
Our experiments on the Human Actions and UCF-101 datasets demonstrate that our MDP-based model is more memory efficient and improves the video quality both in terms of the new and established metrics.
\end{abstract}

%
%
\section{Introduction}

Video synthesis is a very challenging problem~\cite{IVGAN, VideoFlowGAN, TGAN, MoCoGAN, VideoGAN}, arguably even more challenging than the already difficult image generation task \cite{GAN,AutoEncodingVB,ImprovedGAN}.
The temporal dimension of the data introduces an additional mode of variation, since feasible motions are dependent on the object category and the scene appearance.
Consequently, the evaluation of video synthesis methods should account not only for the quality of individual frames but also for their temporal coherence, motion realism, and diversity.

In this work, we take a closer look at the temporal quality of \emph{unconditional} video generators, represented by the state-of-the-art MoCoGAN approach~\cite{MoCoGAN}.
Note that this subcategory of video generation is different from future frame prediction~\cite{SAVP, PredBeyond}, which takes a number of initial frames as input.
We only rely on the training data as input instead.\footnote{Note that the model is still conditioned on the particular training data distribution, hence not truly ``unconditional''.
Still, we adhere to the common terminology used in the literature.}

\begin{figure}[!t]
\centering

\def\svgwidth{\columnwidth}
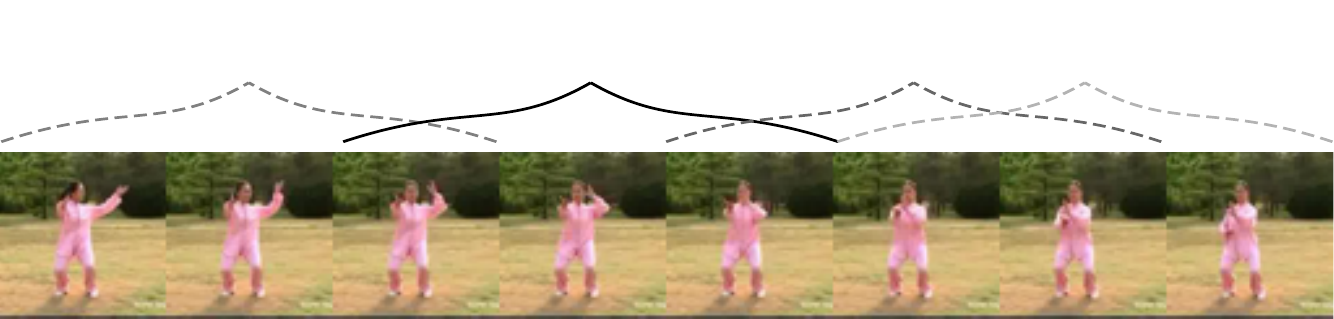

\includegraphics[width=\linewidth]{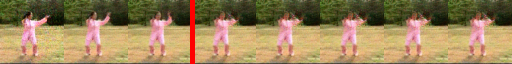} \\
\includegraphics[width=\linewidth]{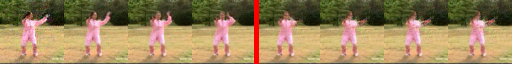} \\
\includegraphics[width=\linewidth]{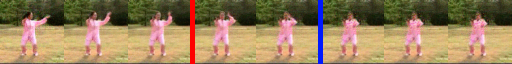} \\

\caption{\textbf{Problem illustration on a Tai Chi sequence.} Every \nth{6} frame is shown. \textbf{Top row}: The ground truth video is a non-repetitive action sequence. \textbf{Second row}: Even when trained only on one video, MoCoGAN~\cite{MoCoGAN} can only reproduce the sequence until the training length, marked by the red boundary, and the motion freezes thereafter. \textbf{Third row}: Increasing the training length comes at increased memory costs and only delays the freezing. \textbf{Last row}: Our MDP approach uses shorter training sequences yet extends the movement duration, indicated by the blue boundary.}
\label{fig:FreezingVideosTaiChi}
\vspace{-0.5em}
\end{figure}

We find that the common training strategy of sampling a fixed-length video subsequence at training time often leads to degenerate solutions.
As illustrated in \cref{fig:FreezingVideosTaiChi}, the MoCoGAN model exhibits temporal artifacts as soon as the video sequence length at inference time exceeds the length of the temporal window at training time.
We establish two common types of such artifacts.
If the model continues to predict the last frame without change, we refer to that as \emph{freezing}.
On the other hand, \emph{looping} occurs when the exact subsequence of frames is continually repeated.

To address these limitations, we make two main contributions.
First, to tackle the detrimental effect of fixed-length video training, we reformulate video generation as a \emph{Markov Decision Process} (MDP).
This reformulation allows approximating an infinite forecast horizon in order to optimize every generated frame \wrt to its long-term effect on future frames.
One benefit of our MDP formulation is that it is model-agnostic.
We evaluate it by applying it to the state-of-the-art MoCoGAN \cite{MoCoGAN}, which requires only a minor modification of the original design and does not significantly increase the model capacity.
Second, we propose a \emph{family of evaluation metrics} to detect and measure the temporal artifacts.
Our new metrics are model-free, simple to implement, and offer an easy interpretation.
In contrast to the Inception Score (IS)~\cite{ImprovedGAN} or the recent Fr\'echet Video Distance (FVD)~\cite{FrechetVD}, the proposed metrics do not require model pre-training and, hence, do not build upon a data-sensitive prior.
Our experiments show that our MDP-based formulation leads to a consistent improvement of the video quality, both in terms of the artifact mitigation as well as on the more common metrics, the IS and FVD scores.

%
%
\pagestyle{plain}
\section{Related Work}
\label{sec:related_work}
Video generation models can be divided into two main categories: \emph{conditional} and \emph{unconditional}.
Exemplified by the task of future frame prediction, conditional models historically preceded the latter and some of their features lend themselves to unconditional prediction.
Therefore, we first give a brief overview of conditional approaches.

\myparagraph{Conditional video generation}. One of the first network-based models for motion dynamics used a temporal extension of Restricted Boltzmann Machines (RBMs)~\cite{SutskeverH07, HumanMotionRBM} with a focus on resolving the intractable inference~\cite{RTRBM}.
The increasing volume of video data for deep learning shifted the attention to learning suitable representations and enabling some control over the generated frames~\cite{HolisticControl}.
Srivastava~\etal~\cite{UnsupervisedVideoRepresentation} show that unsupervised sequence-to-sequence pre-training with LSTMs~\cite{HochreiterS97} enhances the performance on the supervised frame prediction task.
Patch-based quantization of the output space~\cite{VideoPredictionBaseline} or predicting pixel motion~\cite{PhysicalInteraction, LiuYTLA17} can improve the frame appearance at larger resolutions.
In contrast, Kalchbrenner~\etal~\cite{VideoPixelNetwork} predict pixel-wise intensities and extend the context model of PixelCNNs~\cite{OordKEKVG16} to the temporal domain.
A coarse-to-fine strategy allows to decouple the structure from the appearance~\cite{VillegasYHLL17,LongTermFuture}, or dedicate individual stages of a pipeline to multiple scales~\cite{PredBeyond}.

The frames of a distant future cannot be extrapolated deterministically due to the stochastic nature of the problem~\cite{BabaeizadehFECL18,SAVP,Xue0BF16} (\ie there are multiple feasible futures for a given initial frame).
In practice, this manifests itself in frame blurring --  a gradual loss of details in the frame. 
To alleviate this effect, Mathieu~\etal~\cite{PredBeyond} used an adversarial loss~\cite{GAN}.
Liang~\etal~\cite{LiangLDX17} further show that adversarial learning of the pixel flows leads to better generalisation.

\myparagraph{Unconditional video generation}.
These more recent methods are based on the GAN framework~\cite{GAN} and incorporate some of the insights from their conditional counterparts.
For example, Vondrick \etal~\cite{VideoGAN} decouple the active foreground from a static background by using an architecture with two parallel generator streams.
Saito \etal~\cite{TGAN} use two generators to disentangle the video representation into distinct temporal and spatial domains.
Following~\cite{VillegasYHLL17}, the state-of-the-art MoCoGAN of Tulyakov \etal~\cite{MoCoGAN} decomposes the latent representation into content and motion parts for finer control over the generated scene.
In addition, the discriminator in the MoCoGAN model is separated into image and video modules.
While the image module targets the visual quality of individual frames, the focus of the video discriminator is the temporal coherence.

\myparagraph{Evaluating unconditional video generators}.
Borrowed from the image generation literature~\cite{ImprovedGAN}, the Inception Score (IS) has become one of the established metrics for quality assessment in videos~\cite{TGAN,MoCoGAN,VideoGAN}.
IS incorporates the entropy of the class distributions obtained from a separately trained classifier.
Therefore, it is only meaningful if the training data distribution of the classifier matches the one on which it will be evaluated later.
Following~\cite{HeuselRUNH17}, Unterthiner~\etal~\cite{FrechetVD} recently proposed the Fr\'echet Video Distance (FVD) that compares the distributions of \emph{feature embeddings} of real and generated data.

However, these metrics provide only a holistic measure of the video quality and do not allow for a detailed assessment of its individual properties.
One of the desirable qualitative traits of video generators is their \emph{ability to produce realistic videos of arbitrary length}.
Yet, the established experimental protocol evaluates only on video sequences of a fixed length.
Indeed, some previous work~\cite{TGAN,VideoGAN} is even tailored to a pre-defined video length, both at training and at inference time.

%
%
\section{MDP for Video Generation}
\label{sec:Model}

To motivate MDP for video generation, we first review MoCoGAN \cite{MoCoGAN} and discuss its limitations.
After a short presentation of the MDP formalism (\cf \cite{RLBook} for a comprehensive introduction),
we then integrate MDP into MoCoGAN to incorporate knowledge of the infinite-time horizon into the generative process.

\begin{figure*}[t]
\subcaptionbox{\label{fig:MoCoGAN}}{
\hspace{1em}
\def\svgwidth{0.25\linewidth}
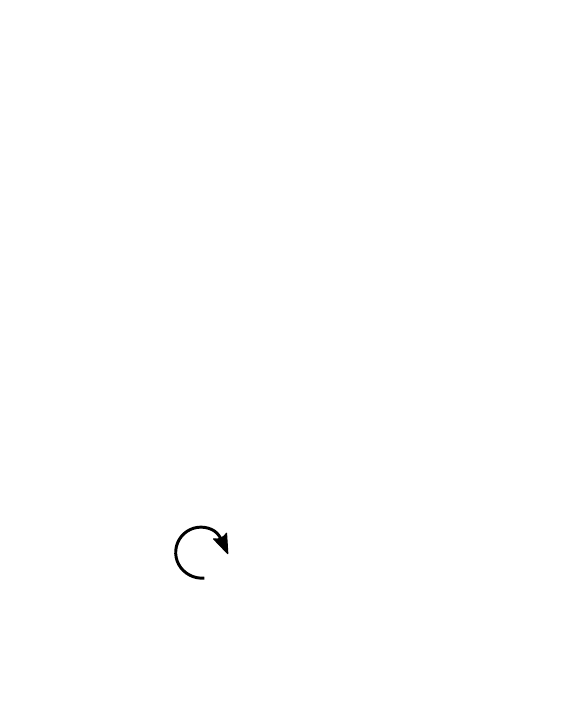
\hspace{2em}
}
\subcaptionbox{\label{fig:OurApproach}}{
\hspace{-6em}
\raisebox{2.4em}{
\def\svgwidth{0.63\linewidth}
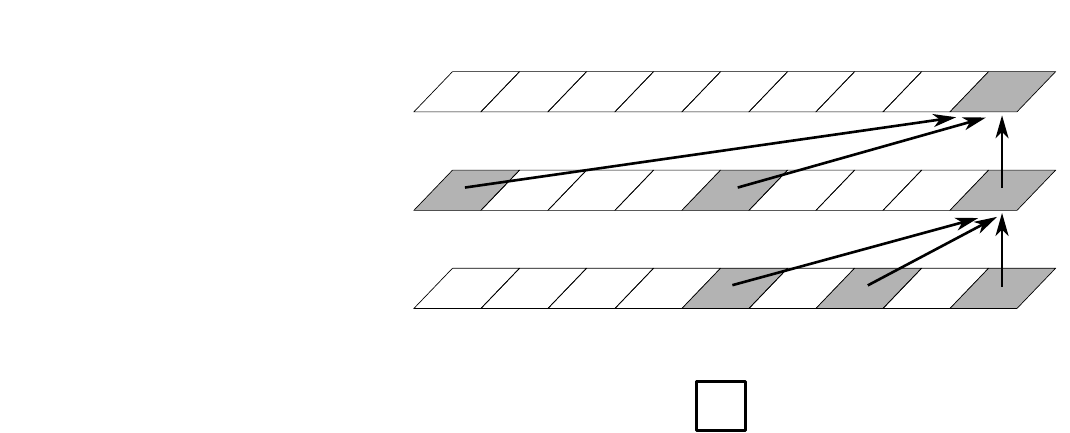
}
}
\vspace{-2mm}
\caption{\textbf{The original MoCoGAN architecture} \subref{fig:MoCoGAN} and our \textbf{proposed modification} of $D_V$ for modeling the MDP \subref{fig:OurApproach}.
Our MDP re-formulation follows the TCN design \cite{TCN}: a sequence of 3D-convolutional layers with layer-specific dilations and strides.
The input to the next convolutional layer is the output of the previous one.
The last layer produces the immediate rewards $\{r_t\}_{1 \leq t \leq K}$ and the $\{Q_t\}_{1 \leq t \leq K}$, \ie the $Q$-values are produced by the same network, $D_V$.
}
\end{figure*}

\subsection{Preliminaries}
\label{sec:mdp_video}

\paragraph{MoCoGAN.}
Figure~\ref{fig:MoCoGAN} illustrates the main components of MoCoGAN: the generator, the image discriminator, and the video discriminator.
At every timestep, the stochastic generator $G$ emits one frame $x_t$ and maintains a recurrent state $h_t$ perturbed by random noise.
The image discriminator $D_I$ provides feedback for a single image; the video discriminator $D_V$ evaluates a contiguous subsequence of frames $\mathbf{x}_t$ of a pre-defined length $\lvert \mathbf{x}_t \rvert = K$.
The training objective is specified by the familiar max-min game
\begin{equation}
\max_{G}\min_{D_I, D_V} \: \underset{x_t, \mathbf{x}_t}{\mathbb{E}} \Big[\mathcal{L}_I (x_t^\text{real}, x_t^\text{fake}) + \mathcal{L}_V(\mathbf{x}_t^\text{real}, \mathbf{x}_t^\text{fake}) \Big],
\begin{aligned}
\label{eq:mocogan_objective}
\end{aligned}
\end{equation}
where $x_t^\text{real}$ and $\mathbf{x}_t^\text{real}$ are samples from the training data, the generator provides $x_t^\text{fake}$ and $\mathbf{x}_t^\text{fake}$, and $\mathcal{L}_I$ and $\mathcal{L}_V$ are defined by the scalar scores of $D_I$ and $D_V$ \cite{GAN,MoCoGAN}.

We find that MoCoGAN's samples exhibit looping and freezing patterns (see \cref{sec:mocogan} for results and analyses).
The intuitive reason comes from the specifics of training:
to save memory, the training samples contain only \emph{subsequences} of the complete video.
As a result, the gradient signal from the video discriminator is unaware of the frames following the subsequence.
The predefined length of the subsequence ultimately determines the maximum length of a sample with a non-repeating pattern.\footnote{To verify this, we also trained the MoCoGAN model on longer subsequences and found the breaking point to occur at a correspondingly later timestep.}

\myparagraph{MDP.}
In an MDP defined by the tuple $\left(S, A, T, \pi, r\right)$, the \textit{agent} interacts with the environment by performing \emph{actions}, $a_t \in A$, based on the current state, $s_t \in S$.
The environment specifies the outcome of the action by returning a \emph{reward}, $r(s_t, a_t)$, and the next state, $s_{t+1} = T(s_t, a_t)$.
The goal of the agent is to find the optimal policy $\pi^\ast: S \rightarrow A$, maximizing the discounted cumulative reward
\begin{equation}
\pi^\ast = \argmax_{\pi} \sum_{t=0}^{\infty} \gamma^t r(s_t, a_t), \quad a_t \sim \pi(s_t),
\label{eq:InfiniteHorizonRewards}
\end{equation}
where $\gamma \in (0, 1)$ is the \emph{discount factor} to ensure the convergence of the sum.

In the context of an MDP the generator $G$ plays the role of the agent's policy.
The frames predicted by $G$ are the actions.
The hidden recurrent state $h_t$ becomes the agent's state $s_t$.
The additive noise at every timestep determines the transition function $T$.
A frame incurs a reward $r_t$ as the score provided by the discriminators.
Due to the deterministic mapping $s_t \rightarrow a_t$, the MoCoGAN's $G$ corresponds to a deterministic policy~\cite{SilverLHDWR14} (\ie the sampling in \cref{eq:InfiniteHorizonRewards} becomes an equality).
The optimization task for the agent is a search for the optimal policy $\pi^\ast$:
\begin{equation}
\max_\pi \: r(s_t, a_t) + \underset{a = \pi(s_t)}{\mathbb{E}} \Bigg[ \sum_{i=t+1}^\infty \gamma^{i-t}  r(s_i, a) \Bigg].
\label{eq:RewardOptEquation}
\end{equation}
Observe that the MoCoGAN objective for $D_V$ is equivalent to only the first term of \cref{eq:RewardOptEquation}, the immediate reward, since the $D_V$ computes only a single score for a given video sample.
In contrast, we also consider the future rewards, \ie the second term of \cref{eq:RewardOptEquation}.
To this end, we decompose the score of the video generator into immediate rewards associated with individual frames.
We then learn a utility Q-function approximating the expected cumulative reward, $\mathbb{E} \big[ \sum_t \gamma^t r_t \big]$.
Its definition is also known as Bellman's optimality principle:
\begin{equation}
Q(s_t, a_t) = r(s_t, a_t) + \underset{a = \pi(s_t)}{\max} Q(s_{t+1}, a).
\label{eq:BellmanEquation}
\end{equation}
By training the generator to maximize the Q-function instead of just the immediate reward, we arrive at an approximate solution of \cref{eq:RewardOptEquation}.
In the next section, we detail how MoCoGAN can be extended to this setup.

\subsection{Integrating MDP into MoCoGAN}
\label{sec:MDPGAN}

We need the model implementing the MDP to comply with two requirements:
\begin{compactenum}[(a)]
\item The \emph{Markov property} needs to be fulfilled, \ie the next state $s_{t+1}$ given the previous state $s_t$ is conditionally independent from the past history $s_{i<t}$.
\item By \emph{causality}, the immediate reward $r_t$ is a function of the current state $s_t$ and the action $a_t$ and incorporates no knowledge about future actions.
\end{compactenum}
\bigskip
The MoCoGAN generator already satisfies the Markov property using a parametrized RNN mapping from the current state to the next.
However, the video discriminator has to be modified to satisfy the second requirement.
This modification is straightforward to implement and leads to a variant of the Temporal Convolutional Network (TCN)~\cite{TCN}.

\Cref{fig:OurApproach} gives an overview of the proposed MDP-extension for the video discriminator.
The key property of this design is that the $t^\text{th}$ output -- a scalar -- corresponds to a temporal receptive field of the frames up to the $t^\text{th}$ timestep.
In this way the immediate reward will capture only the relevant motion history.
Fortunately, adapting the MoCoGAN video discriminator to this architecture is straightforward (\cf supplemental material for more details).

To implement \cref{eq:BellmanEquation}, alongside $r_t$ we also predict another time-dependent scalar, the Q-value.
As discussed in \cref{sec:mdp_video}, the purpose of the Q-value is to approximate the expected cumulative reward, $\mathbb{E} \big[ \sum_{t=0}^\infty \gamma^t  r(s_t, a_t) \big]$.
We use the squared difference loss, defined for each timestep by
\begin{equation}
\mathcal{L}_{Q,t} = \norm{\frac{1}{K-t+1} \sum\limits_{i=t}^K \gamma^{i-t} r_i - Q_t}^{2}_{2}, \: 1 \leq t \leq  K,
\label{eq:TCNApproximation}
\end{equation}
where $\gamma \in (0, 1)$ is the discounting factor specifying the lookahead span: larger values encourage the Q-value to account for the future outcome far ahead;
low values focus the Q-value on the immediate effect of the current frames.

Our TCN-based $D_V$ ensures that the parameters for predicting $Q_t$ are now \emph{shared} for all $t$.
As a result, \cref{eq:TCNApproximation} forces even the last $Q_K$ to incorporate knowledge of rewards \emph{beyond} the temporal window of size $K$.
Hence, by maximizing $Q_K$, the generator will implicitly maximize the rewards for $t>K$.
Contrast this to the original $D_V$ producing a single score for the complete $K$-frame sequence: due to lack of causality, the generator is ``unaware'' that at inference time the requested video length may exceed $K$.

Note that the definition in \cref{eq:TCNApproximation} is confined to a limited time window of length $K$ to ensure that the memory consumption remains manageable.
Now, our task is to train the generator by maximizing the Q-value incorporating the long-term effects of individual predictions.
However, since we keep $K$ fixed, each consecutive $Q_t$ in \cref{eq:TCNApproximation} will be optimized \wrt to the sum containing one term fewer.
That is, $Q_1$ will approximate a sum of $K$ immediate rewards, $Q_2$ a sum of $K-1$ terms, \etc.
As a result, $Q_1$ incorporates the effect of the $1^\text{st}$ frame on $K-1$ future frames, whereas $Q_{K-1}$ will only observe the influence of the ${(K-1)}^\text{th}$ frame on the last prediction.
It is therefore evident that the Q-values are not equally informative for modeling the long-term dependencies as supervision to the generator.

To reflect this observation in our training, we introduce an additional discounting factor $\beta \in [0, 1]$ that shifts the weight of the long-term supervision to the first frames, but offsets the reliance on the Q-value for the last predictions.
Concretely, the new term in the generator loss is
\begin{equation}
\mathcal{L}_{T} = \sum\limits_{t=1}^K \beta^{t} Q_t.
\label{eq:GenLossT}
\end{equation}
To summarize, extending the original MoCoGAN training objective (\Eq~\ref{eq:mocogan_objective}) into our MDP-based GAN yields
\begin{subequations}
\begin{equation}\label{eq:TCNlossD}
\begin{aligned}
\min_{D_I, D_V} \underset{x_t, \mathbf{x}_t}{\mathbb{E}} & \bigg[ \mathcal{L}_I (x_t^\text{real}, x_t^\text{fake}) + \mathcal{L}_V(\mathbf{x}_t^\text{real}, \mathbf{x}_t^\text{fake})
\\
& + \frac{1}{K} \sum_{t=1}^K \big( \mathcal{L}_{Q,t}(\mathbf{x}_t^\text{real}) + \mathcal{L}_{Q,t}(\mathbf{x}_t^\text{fake}) \big) \bigg],
\end{aligned}
\end{equation}
\begin{equation}\label{eq:TCNlossG}
\begin{aligned}
\max_G \underset{x_t, \mathbf{x}_t}{\mathbb{E}} \big[ & \mathcal{L}_I (x_t^\text{fake}) + \mathcal{L}_V(\mathbf{x}_t^\text{fake}) + \mathcal{L}_{T} \big].
\end{aligned}
\end{equation}
\end{subequations}
Here, we split the original objective in~\cref{eq:mocogan_objective} into the discriminator- and generator-specific losses for illustrative purposes although the joint nature of the max-min optimization problem remains.
Following standard practice~\cite{GAN}, we optimize the new objective by alternately updating the discriminators using \cref{eq:TCNlossD} and the generator using \cref{eq:TCNlossG}.

%
%
\section{Quantifying Temporal Diversity}
\label{sec:Metrics}

Motivated by our observation of the looping and freezing artifacts~(see \cref{fig:FreezingVideosTaiChi}), we propose an interpretable way to quantify the temporal diversity of the video.
Here, our assumption is that realistic videos comprise a predominantly unique sequence of frames.
The idea then is to compare the predicted frame to the preceding ones:
if there is a match, this indicates a re-occurring pattern in the sequence.

Let $X=(x_t)_{t=1..N}$ be a sequence of frames predicted by the model.
Our diversity measure relies on a distance function of choice between arbitrary frames $d(x_i, x_j)$ as
\begin{equation}
\label{eq:metric}
\text{\textit{t-}}d = \frac{1}{N} \sum_{i=2}^N \min_{j < i} d(x_i, x_j),
\end{equation}
where we use prefix ``\textit{t-}'' for disambiguation.
\cref{eq:metric} essentially finds the most similar preceding frame and averages the distance over all such pairs in the sequence.
The obvious dual of this metric is to replace the distance function $d(\cdot, \cdot)$ in \cref{eq:metric} with a similarity measure $s(\cdot, \cdot)$ and substitute the $\min$ for the $\max$ operation.
In this work, we use two instantiations of \cref{eq:metric}:
the \emph{\tssim} employs the structural similarity (SSIM)~\cite{MetricSSIM} in the distance function $\text{DSSIM} = \tfrac{1}{2}(1 - \text{SSIM})$;
\emph{\tpsnr} utilizes the peak signal-to-noise ratio (PSNR) as a similarity measure.
Hence, higher \tssim and lower \tpsnr indicate higher diversity of frames within a sequence.
We show next that despite its apparent simplicity, our proposed metric effectively captures deficiencies in frame diversity.

%
%
\section{Experiments}

\begin{figure*}[t]
\begin{floatrow}
\capbtabbox[1.2\linewidth]{%
\small
\begin{tabular}{@{}cl@{\hskip 1.2em}c@{\hskip 1.2em}c@{\hskip 1.2em}c@{\hskip 1.2em}c@{}}
\toprule
& Configuration & IS $\uparrow$ & FVD $\downarrow$ & \tssim $\uparrow$ & \tpsnr $\downarrow$ \\ \midrule

\parbox[t]{2mm}{\multirow{4}{*}{\rotatebox[origin=c]{90}{Tai Chi}}} & Original & 1.63 {\footnotesize$\pm$ 0.05} & \bfseries 115.3 {\footnotesize$\pm$ 6.9} & \bfseries 0.013 & \bfseries 36.50 \\
& Looping-FWD & \bfseries 2.03 {\footnotesize$\pm$ 0.03} & 336.7 {\footnotesize$\pm$ 13.5} & 0.0062 & $\infty$ \\
& Looping-BWD & 1.69 {\footnotesize$\pm$ 0.03} & 541.7 {\footnotesize$\pm$ 19.4} & 0.0062 & $\infty$ \\
& Freezing & 1.55 {\footnotesize$\pm$ 0.05} & 254.4 {\footnotesize$\pm$ 15.5} & 0.0062 & $\infty$ \\ \midrule

\parbox[t]{2mm}{\multirow{5}{*}{\rotatebox[origin=c]{90}{UCF-101}}} & Original & \bfseries 40.74 {\footnotesize$\pm$ 0.20} & 472.8 {\footnotesize$\pm$ 18.5} & 0.073 & 27.10 \\
& Original + $\epsilon$ & 36.69 {\footnotesize$\pm$ 0.23} & \bfseries 444.8 {\footnotesize$\pm$ 17.2} & \bfseries 0.107 & \bfseries 25.44 \\
& Looping-FWD & 38.59 {\footnotesize$\pm$ 0.22} & 597.2 {\footnotesize$\pm$ 13.5} & 0.034 & $\infty$ \\
& Looping-BWD & 35.16 {\footnotesize$\pm$ 0.78} & 737.7 {\footnotesize$\pm$ 40.0} & 0.034 & $\infty$ \\
& Freezing & 32.45 {\footnotesize$\pm$ 0.22} & 667.3 {\footnotesize$\pm$ 17.8} & 0.034 & $\infty$ \\ \bottomrule
\end{tabular}
}{%
    \caption{\textbf{Comparison of IS, FVD, \tssim, and \tpsnr metrics} for ground-truth videos and videos with purposely crafted artifacts. The Gaussian noise $\epsilon$ is drawn from $\mathcal{N}(\mu = 0, \sigma^2 = 0.03)$.}
    \label{table:MetricsLimitations}
}%
\hspace{-1em}%
\ffigbox[0.84\linewidth]{%
  \includegraphics[width=\linewidth]{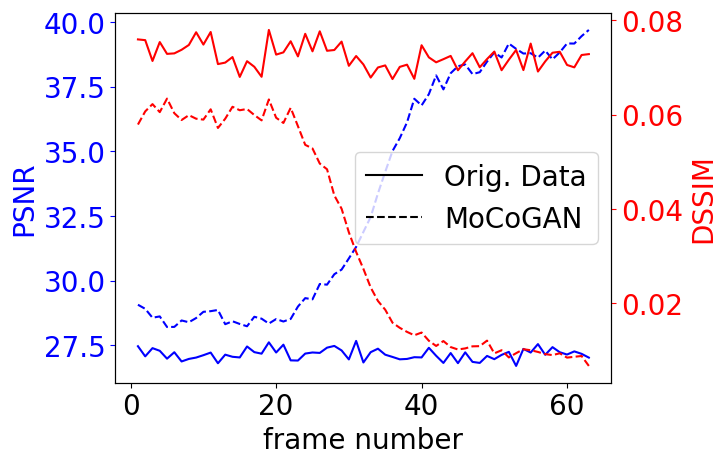}
}{%
  \caption{\textbf{\tpsnr and \tssim decomposed as functions of time.} In contrast to the ground truth, the diversity of the MoCoGAN samples vanishes with time.}%
  \label{fig:metric_plots}
}
\end{floatrow}
\vspace{-0.5em}
\end{figure*}

\subsection{Datasets}
\label{sec:datasets}

Following the established evaluation protocol from previous studies \cite{TGAN, MoCoGAN}, we use the following benchmarks:

\begin{compactenum}[(1)]
\item \textbf{Human actions}~\cite{HumanActions}:
The dataset contains 81 videos of 9 people performing 9 actions, \eg walking, jumping, etc.
All videos are extracted with $25$ fps and down-scaled to $64\times64$ pixels.
We also add a flipped copy of each video sequence to the training set.
Following Tulyakov \etal \cite{MoCoGAN} we used only 4 action classes, which amounts to 72 videos for training in total.

\item \textbf{UCF-101}~\cite{UCF-101}: This dataset consists of 13\,220 videos with 101 classes of human actions grouped into 5 categories: human-object and human-human interaction, body motion, playing musical instruments, and sports.
This dataset is challenging due to a high diversity of scenes, motion dynamics, and viewpoint changes.

\item \textbf{Tai-Chi}:
The dataset contains 72 Tai Chi videos taken from the UCF-101 dataset.\footnote{Note that the Tai Chi subset used in the evaluation of MoCoGAN~\cite{MoCoGAN} is not publicly available and could not be obtained due to licensing restrictions.}
All videos are centered on the performer and downscaled to $64\times64$ pixels.
We use this dataset for our ablation studies as it has moderate complexity, yet represents real-world motion.
\end{compactenum}

\subsection{Overview}
\label{sec:ExpOverview}

We first verify that \tpsnr and \tssim effectively quantify the temporal artifacts.
We then employ these metrics to analyze the MoCoGAN model \cite{MoCoGAN} \wrt these artifacts.
Next, we study the effect of the time-horizon hyperparameters, $\gamma$ and $\beta$, of our MDP approach.
Finally, we validate our approach on the Human Actions dataset and on the more challenging UCF-101 dataset.
We compare our model to TGAN~\cite{TGAN} and MoCoGAN, where we find a consistent improvement of the temporal diversity over the baseline.

We compute the IS following Saito \etal~\cite{TGAN}, who trained the C3D network~\cite{C3D} on the Sports-1M dataset \cite{Sport1M} and then further finetuned on UCF-101 \cite{UCF-101}.
For FVD we use the original implementation by Unterthiner \etal~\cite{FrechetVD}.
To manage computational time, we calculate the FVD for the first 16 frames, sampled from 256 videos, and derive the FVD mean and variance from 4 trials, similar to IS.

\begin{table*}[!t]
\centering
\begin{tabularx}{\textwidth}{@{}Xccccccc@{}}
\toprule
\multirow{2}{*}{Metric} & \multirow{2}{*}{\begin{tabular}{c}\makecell{Tai Chi}\end{tabular}} & \multirow{2}{*}{\begin{tabular}{c}\makecell{MoCoGAN\\$K=16$}\end{tabular}} & \multicolumn{5}{c}{MDP model} \\ \cmidrule(lr){4-8}
&
&
& \makecell{$\gamma=0.0$ \\ $\beta=0.0$}      
& \makecell{$\gamma=0.7$ \\ $\beta=0.7$}      
& \makecell{$\gamma=0.7$ \\ $\beta=0.9$}
& \makecell{$\gamma=0.9$ \\ $\beta=0.7$}    
& \makecell{$\gamma=0.9$ \\ $\beta=0.9$} \\ \midrule

\makecell[l]{IS $\uparrow$} & 1.63 $\pm$ 0.05 & 4.49 $\pm$ 0.04 & \bfseries 4.52 $\pm$ 0.05 &  4.15 $\pm$ 0.04 & 4.24 $\pm$ 0.06 & 3.92 $\pm$ 0.07 & 3.99 $\pm$ 0.04 \\

\makecell[l]{FVD $\downarrow$} & 118 $\pm$ 5 & 828 $\pm$ 38 & 1108 $\pm$ 50 & 787 $\pm$ 10 & 782 $\pm$ 40 & \bfseries 744 $\pm$ 40 & 809 $\pm$ 22 \\

\makecell[l]{\tssim $\uparrow$} & 0.0135 & 0.0031 & 0.0031 & 0.0024 & \bfseries 0.0037 & 0.0035 & 0.0035 \\

\makecell[l]{\tpsnr $\downarrow$} & 36.48 & 45.37 & 57.34 & 50.16 & 44.87 & \bfseries 44.39 & 45.06 \\ \bottomrule

\end{tabularx}
\caption{\textbf{Results of the ablation study of the MDP approach on the Tai Chi dataset.} Our MDP configurations assume a selection of hyperparameters $\beta$ and $\gamma$. For comparison, we include the results from the MoCoGAN baseline. By leveraging the long-term rewards, our MDP model improves the temporal diversity (\tpsnr and \tssim) and FVD scores at the cost of a slight drop in IS.}
\label{table:Ablation}
\end{table*}

\subsection{Metric evaluation}
\label{sec:metric_eval}

We design a set of proof-of-concept experiments to study the properties of the newly introduced \tpsnr and \tssim.
Concretely, we synthesize the looping and freezing patterns in the ground-truth videos from UCF-101 and Tai Chi.
We construct 16 frames by sampling 8 frames directly from the dataset and completing the sequence with an artifact counterpart.
\texttt{Looping-FWD} contains a repeating subsequence from the original video (\texttt{Original}), whereas \texttt{Looping-BWD} reverses the frame order.
The size of the re-occurring subsequence in \texttt{Freezing} is one.
To put the results in context, we also compare to the mainstream IS as well as the recent FVD scores and study the robustness of all metrics to additive Gaussian noise $\epsilon \sim \mathcal{N}(\mu, \sigma^2)$.
The results are summarized in Table~\ref{table:MetricsLimitations}.

We observe that \tpsnr and \tssim correlate well with the more sophisticated IS and FVD.
Recall that both IS and FVD require training a network on videos of fixed length, hence
\emph{(i)} can be computed only for short-length videos, due to GPU constraints;
\emph{(ii)} may be misleading (\eg Tai Chi results in Table~\ref{table:MetricsLimitations}) when the training data for the inception network is different from the evaluated data.
By contrast, \tpsnr and \tssim prove to be faithful in quantifying the artifacts we study, as they are data-agnostic and accommodate videos of arbitrary length.
However, our metrics are permutation invariant, do not assess the quality of the frames themselves, and are not robust to random noise.
Hence we stress their \emph{complementary role} to IS and FVD as a measure of the \emph{overall} video quality.

\begin{figure}[t]
\centering
\includegraphics[width=\linewidth]{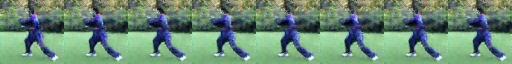} \\
\includegraphics[width=\linewidth]{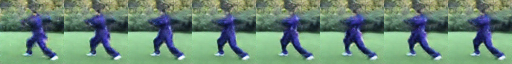}\\
\caption{\textbf{Tai Chi comparison} between MoCoGAN (\textbf{top row}) exhibiting the freezing artifact, and our MDP model (\textbf{bottom row}) generating perceivable motion (\eg torso).}
\label{fig:AblationSamples}
\end{figure}

\subsection{MoCoGAN: a case study}
\label{sec:mocogan}

Here, we study the temporal diversity of the MoCoGAN model~\cite{MoCoGAN} using our \tpsnr and \tssim scores.

We train MoCoGAN\footnote{We use the publicly available code provided by the MoCoGAN authors at \url{https://github.com/sergeytulyakov/mocogan}.} on the UCF-101 dataset with temporal windows of size $K = 16$, and apply our temporal metrics to the samples from the generator.
To enable a more detailed view of the temporal dynamics, we inspect the video samples as a function of time in \cref{fig:metric_plots} by plotting the values of the summands in \cref{eq:metric} for each timestep.
To rule out the possibility of any degenerate phenomena in the original data, we also plot the corresponding curves of the ground-truth sequences alongside.
This clearly shows that MoCoGAN exhibits a vanishing diversity of video frames -- a pattern that is not found in the training data.

\subsection{MDP approach: an ablation study}
\label{sec:Ablation}
Here, we perform an ablation study of our MDP approach by varying the time-horizon hyperparameters, $\gamma$ and $\beta$, introduced in \cref{sec:MDPGAN}.
Recall that $\gamma$ controls the timespan of the future predictions modeled by the Q-value: lower values imply a shorter time horizon, whereas higher values encourage the model to learn long-term dependencies.
Parameter $\beta$, on the other hand, specifies how accounting for the long-term effect is distributed over the timesteps.
High values specify equal distribution; lower values force the model to encode the long-term effects more in the earlier than in the later timesteps.
As a boundary case, we also consider $\beta=0$ and $\gamma=0$ to gage the effect of the architecture change in the video discriminator (TCN), which is needed to implement reward causality (\cf \cref{sec:MDPGAN}).
As quantitative measures, we use the Inception Score (IS)~\cite{ImprovedGAN}, the Fr\'echet Video Distance (FVD)~\cite{FrechetVD}, as well our temporal metrics, \tssim and \tpsnr, introduced in \cref{sec:Metrics}.

The results in \cref{table:Ablation} show that by leveraging the increasing values of the time-horizon hyperparameters, our model clearly improves the temporal diversity in terms of \tpsnr and \tssim.
Moreover, we also observe that the TCN baseline ($\gamma=0$, $\beta=0$) performs worse than the original MoCoGAN in terms of temporal diversity.
This is easily understood when considering that the TCN alone does not have any lookahead into the future (\cf \cref{fig:OurApproach}).
However, once we enable taking the future rewards into account by virtue of our MDP formulation, we not only reach but actually surpass the temporal diversity of the baseline MoCoGAN, as expected.

The somewhat inferior IS and FVD scores might be due to their sensitivity to the data prior, as discussed in \cref{sec:metric_eval}.
This hypothesis is also supported by a qualitative comparison between MoCoGAN and our MDP model.
\Cref{fig:AblationSamples} gives one such example; more results can be found in the supplemental material.
While we observe no notable difference in per-frame quality, the motion between consecutive frames from our MDP model is more apparent than the samples from MoCoGAN (\eg, the torso of the performer).

\begin{table*}[t]
\small
\begin{tabularx}{\textwidth}{@{}lccccccccc@{}}
\toprule
\multirow{2}{*}{Model} & \multirow{2}{*}{K} & \multicolumn{4}{c}{Human Actions} & \multicolumn{4}{c}{UCF-101}
  \\ \cmidrule(lr){3-6} \cmidrule(lr){7-10}

&
& \makecell{IS $\uparrow$}      
& \makecell{FVD $\downarrow$}
& \makecell{\tssim $\uparrow$}    
& \makecell{\tpsnr $\downarrow$}

& \makecell{IS $\uparrow$}      
& \makecell{FVD $\downarrow$}
& \makecell{\tssim $\uparrow$}    
& \makecell{\tpsnr $\downarrow$} \\ \midrule

Raw dataset & -- & 3.39 $\pm$ 0.08 & 49 $\pm$ 2 & 0.0815 & 23.35 & 40.80 $\pm$ 0.26 & 452 $\pm$ 49 & 0.0723 & 28.34 \\ \cmidrule{1-10}

 TGAN (Normal) & 16 & 2.90 $\pm$ 0.04 & 977 $\pm$ 31 & -- & -- & 8.11 $\pm$ 0.07 & 1686 $\pm$ 24 & -- & -- \\

 TGAN (SVC) & 16 & \bfseries 3.65 $\pm$ 0.10 & \bfseries 227 $\pm$ 10 & -- & -- & 11.91 $\pm$ 0.21 & 1324 $\pm$ 23 & -- & -- \\

 MoCoGAN & 16 & 3.53 $\pm$ 0.02 & 300 $\pm$ 8 & 0.0259 & 33.76 & 11.15 $\pm$ 0.10 & 1351 $\pm$ 49 & 0.0337 & 33.29 \\

 MoCoGAN-$D_V^+$ & 16 & 3.51 $\pm$ 0.02 & 245 $\pm$ 6 & 0.0243 & 34.79 & 11.48 $\pm$ 0.15 & 1314 $\pm$ 45 & 0.0358 & 33.61 \\ 

 MoCoGAN & 24 & 3.47 $\pm$ 0.02 & 318 $\pm$ 9 & 0.0254 & 35.72 & 10.49 $\pm$ 0.09 & 1352 $\pm$ 49 & 0.0387 & 32.63 \\ \cmidrule{1-10}

 MDP-0 (ours) & 16 & 3.55 $\pm$ 0.03 & 1413 $\pm$ 15 & 0.0559 & 33.31 & 6.16 $\pm$ 0.08 & 2147 $\pm$ 87 & 0.0160 & 47.36 \\

 MDP (ours) & 16 & 3.55 $\pm$ 0.02 & 641 $\pm$ 8 & 0.0604 & 30.12 & 11.86 $\pm$ 0.11 & \bfseries 1277 $\pm$ 56 & 0.0370 & 32.77 \\

 MDP (ours) & 24 & 3.49 $\pm$ 0.03 & 686 $\pm$ 12 & \bfseries 0.0661 & \bfseries 29.39 & \bfseries 12.14 $\pm$ 0.18 & 1293 $\pm$ 58 & \bfseries 0.0454 & \bfseries 31.05 \\ \bottomrule

\end{tabularx}
\caption{\textbf{Comparison of our two MDP models to the state of the art.} Temporal metrics are calculated for $64$ frames. Our MDP model consistently improves the temporal video quality in terms of \tpsnr, \tssim, and IS.
Moreover, it is more memory efficient as it is comparable to MoCoGAN $K=24$ and can produce videos of arbitrary length in contrast to TGAN. Note that since TGAN \cite{TGAN} can only generate videos of 16 frames, we do not compute \tpsnr and \tssim for this model here.}
\label{table:OverallResults}
\end{table*}

\begin{figure}[t]
\centering

\begin{subfigure}{\linewidth}
  \includegraphics[width=\linewidth]{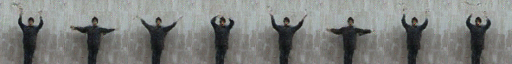} \\
  \includegraphics[width=\linewidth]{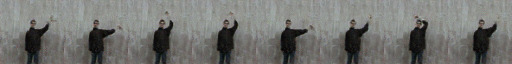}
  \caption{}
  \label{fig:ResultsSamplesHumanAction-a}
\end{subfigure}
\vspace{2mm}

\begin{subfigure}{\linewidth}
  \includegraphics[width=\linewidth]{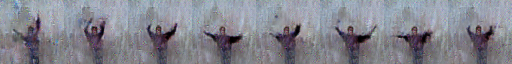} \\
  \includegraphics[width=\linewidth]{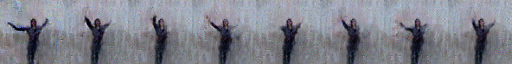}
  \caption{}
  \label{fig:ResultsSamplesHumanAction-b}
\end{subfigure}
\vspace{2mm}

\begin{subfigure}{\linewidth}
  \includegraphics[width=\linewidth]{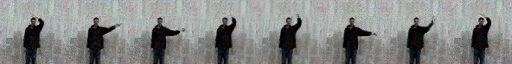} \\
  \includegraphics[width=\linewidth]{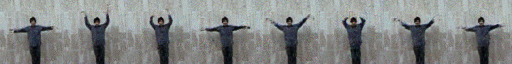}
  \caption{}
  \label{fig:ResultsSamplesHumanAction-c}
\end{subfigure}
\vspace{-2mm}
\caption{\textbf{Random samples on Human Actions.} \subref{fig:ResultsSamplesHumanAction-a} MoCoGAN, \subref{fig:ResultsSamplesHumanAction-b} MDP-0, \subref{fig:ResultsSamplesHumanAction-c} MDP. Disabling MDP leads to poorer video quality in \subref{fig:ResultsSamplesHumanAction-b}, while modelling long-term rewards leads to comparable per-frame quality of the samples from our MDP model \subref{fig:ResultsSamplesHumanAction-c} \wrt MoCoGAN baseline \subref{fig:ResultsSamplesHumanAction-a}, also reflected by IS, yet tangibly higher temporal diversity measured by \tpsnr and \tssim. From the video sequence of $64$ frames, every \nth{8} frame is shown.}
\label{fig:ResultsSamplesHumanActions}
\end{figure}

\subsection{Human Actions and UCF-101}
\label{sec:OverallResults}

We perform further experiments on the Human Actions and the more challenging UCF-101 datasets.\footnote{To ensure a fair comparison, we use the same inception network for IS and FVD and train other methods \cite{TGAN,MoCoGAN} using the authors' implementation (\cf supplemental material for details).}
We select $\gamma = 0.9, \beta = 0.7$ for our MDP model, which provide a good trade-off between the improved \tpsnr, \tssim, FVD and only a slight drop of IS on Tai Chi (\cf \cref{sec:ExpOverview}).
For reference, we train the TCN baseline, \texttt{MDP-0}, by setting $\gamma = 0$ and $\beta = 0$ to decouple the influence of modeling the long-term effects from the changes in the MoCoGAN architecture to comply with reward causality.
We also train our MDP model and MoCoGAN on an extended temporal window $K=24$.
Recall that higher $K$ require more GPU memory, but give the model an advantage, since it observes longer sequences at training time.
Therefore, we aim to mitigate the artifacts while keeping $K$ constant.

The quantitative results are summarized in \cref{table:OverallResults}.
For both the Human Actions and UCF-101 datasets, we observe a consistent improvement of our MDP model in terms of temporal diversity measured by \tpsnr and \tssim.
Moreover, our model also outperforms MoCoGAN in terms of IS on both datasets, as well as FVD on the UCF-101 dataset.
This can be explained by the more varied nature of motion on these datasets compared to the Tai Chi dataset, which makes taking into account future frames more important.
On the Human Actions dataset, the FVD score for our model is inferior to MoCoGAN.
Recall from \cref{sec:ExpOverview}, that for IS and FVD metrics we did not fine-tune the inception classifiers on the Human Actions dataset, which impedes the interpretability of the scores on this dataset.
A visual inspection of the per-frame quality (\cf \cref{fig:ResultsSamplesHumanActions} for examples) reveals no perceptual loss compared to the baseline model.
In contrast, disabling MDP modeling (\texttt{MDP-0}) leads to a clear deterioration in video quality.

\begin{figure}[t]
\centering

\begin{subfigure}{\linewidth}
  \includegraphics[width=\linewidth]{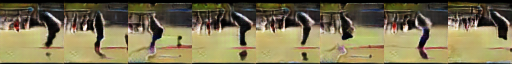} \\
  \includegraphics[width=\linewidth]{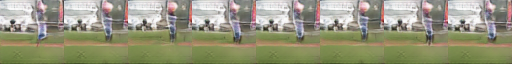}
  \caption{}
  \label{fig:ResultsSamplesUCF-a}
\end{subfigure}
\vspace{2mm}

\begin{subfigure}{\linewidth}
  \includegraphics[width=\linewidth]{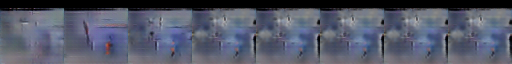} \\
  \includegraphics[width=\linewidth]{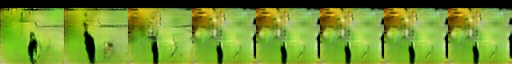}
  \caption{}
  \label{fig:ResultsSamplesUCF-b}
\end{subfigure}
\vspace{2mm}

\begin{subfigure}{\linewidth}
  \includegraphics[width=\linewidth]{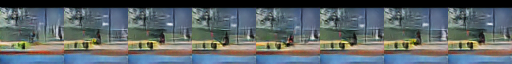} \\
  \includegraphics[width=\linewidth]{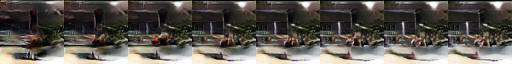}
  \caption{}
  \label{fig:ResultsSamplesUCF-c}
\end{subfigure}
\caption{\textbf{Random samples of the MoCoGAN baseline and MDP models on UCF-101.} \subref{fig:ResultsSamplesUCF-a} MoCoGAN with looping artifact. \subref{fig:ResultsSamplesUCF-b} Our MDP-0 without modeling future rewards exhibits a freezing pattern. \subref{fig:ResultsSamplesUCF-c} Our MDP model. In \subref{fig:ResultsSamplesUCF-c}, while the first sample has some looping, the second does not have temporal artifacts. From the video sequence of $64$ frames, every \nth{8} frame is shown.}
\label{fig:ResultsSamplesUCF}
\end{figure}

On both datasets, our model with $K=16$ is also superior to MoCoGAN with $K=24$ in terms of IS and FVD, and reaches on par performance in terms of \tpsnr and \tssim.
Yet, our MDP-based formulation is significantly more memory efficient, since extending the temporal window at training incurs addition memory costs.
Concretely, at training time the MDP model with $K=16$ consumes roughly $20\%$ more memory than MoCoGAN, whereas setting $K=24$ for the original MoCoGAN incurs a $50\%$ higher memory footprint.
Note that simply increasing the number of parameters of $D_V$ in MoCoGAN is less effective than our proposed MDP approach (see \texttt{MoCoGAN-}$D_V^+$ in Tab.~\ref{sec:OverallResults}).
Also, our MDP model with $K = 24$ improves further over $K=16$ on UCF-101 and regarding the temporal metrics on Human Actions.
A visual inspection of the samples from Human Actions did not reveal any perceptible difference to MoCoGAN or our MDP with $K=16$, despite the inferior IS and FVD scores; we believe this to be an artifact of the evaluation specifics.
The IS score of our MDP model is slightly inferior only to TGAN \cite{TGAN}.
However, TGAN can produce video sequences of only fixed length, whereas our MDP model can generate videos of arbitrary length, owing to the recurrent generator.

\begin{figure}[t]
\centering
\begin{subfigure}{\linewidth}
\begin{tabular}{cc}
  \includegraphics[width=.425\linewidth]{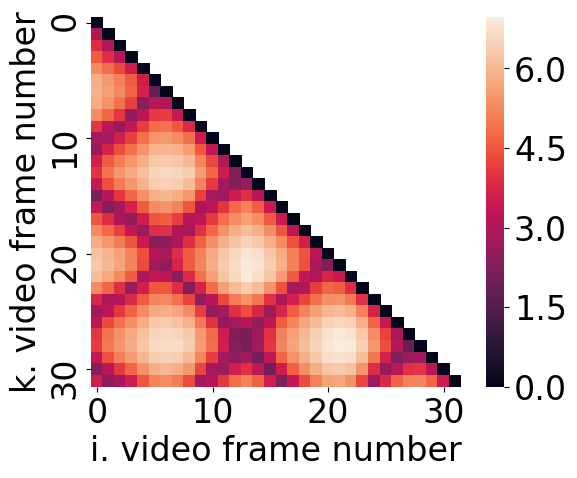} & \includegraphics[width=.425\linewidth]{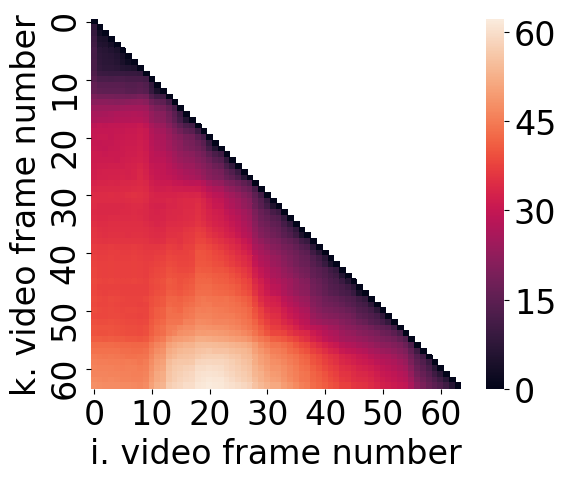}
\end{tabular}
\vspace{-2mm}
\caption{}
\label{fig:ResultsHeatmapHumanActions-a}
\end{subfigure}
\vspace{2mm}

\begin{subfigure}{\linewidth}
\begin{tabular}{cc}
\includegraphics[width=.425\linewidth]{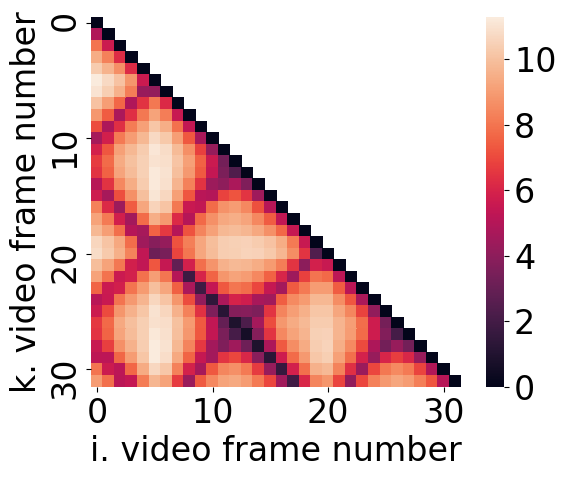} & \includegraphics[width=.425\linewidth]{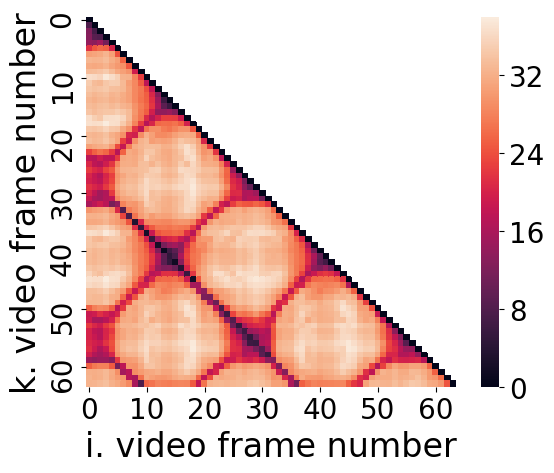}
\end{tabular}
\vspace{-2mm}
\caption{}
\label{fig:ResultsHeatmapHumanActions-b}
\end{subfigure}
\vspace{2mm}

\begin{subfigure}{\linewidth}
\begin{tabular}{cc}
\includegraphics[width=.425\linewidth]{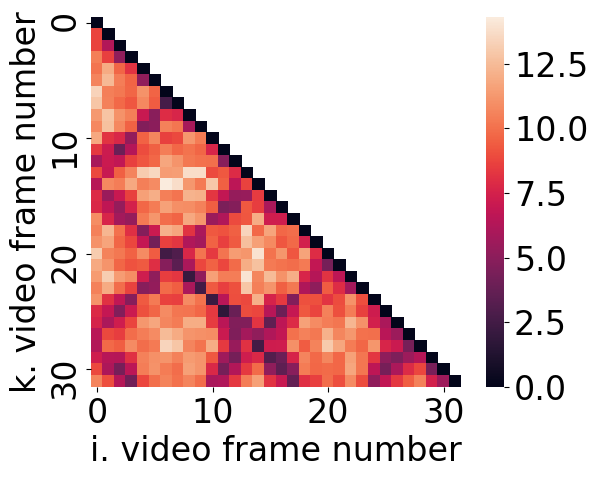} & \includegraphics[width=.425\linewidth]{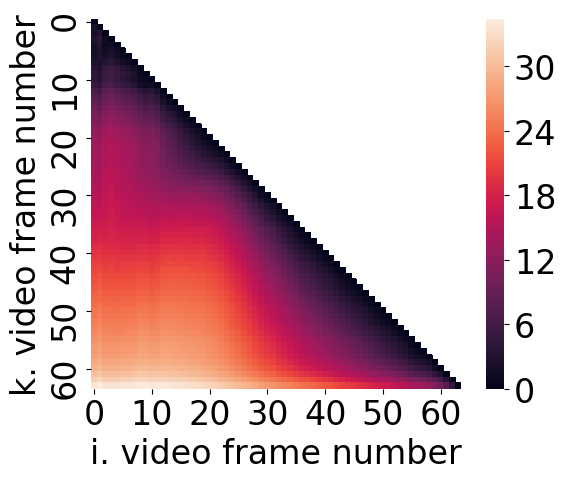}
\end{tabular}
\vspace{-2mm}
\caption{}
\label{fig:ResultsHeatmapHumanActions-c}
\end{subfigure}
\caption{\textbf{Heatmap comparison between ground truth, MoCoGAN, and our MDP models} trained on the Human Actions dataset \textbf{(left)} and UCF-101 \textbf{(right)} (different scales). \subref{fig:ResultsHeatmapHumanActions-a} ground truth, \subref{fig:ResultsHeatmapHumanActions-b} MoCoGAN, \subref{fig:ResultsHeatmapHumanActions-c} MDP ($\gamma=0.9, \beta=0.7$). Our MDP model alleviates the looping artifact on Human Actions, where it can still appear natural. On the more complex UCF-101, our MDP is able to approximate the temporal quality of the ground truth.}
\label{fig:ResultsHeatmapHumanActions}
\end{figure}

The qualitative results in \cref{fig:ResultsSamplesUCF} show that our model can generate complex scenes from UCF-101 that are visually comparable to the MoCoGAN samples.
Similar to our observation on Human Actions, \texttt{MDP-0} produces poorer samples, which asserts the efficacy of the underlying MDP.
Since the interpretation of the UCF-101 results is difficult, we examine a visualization of a pairwise $L_1$-distance between two frames in the video, shown in \cref{fig:ResultsHeatmapHumanActions}.
The distance matrix can be represented as a lower triangular two-dimensional heatmap, owing to the symmetry of $L_1$.
We observe that while MoCoGAN exhibits a looping pattern, our MDP approach tends to preserve the temporal qualities of the ground-truth datasets.
Note that some samples in Human Actions can be naturally periodic (\eg hand-waving), hence, we do not expect our model to dispense with the looping pattern completely.
The overall results suggest that modeling long-term dependencies with an MDP consistently leads to more diverse motion dynamics, which becomes more apparent in increasingly complex scenes.

%
%
\section{Conclusions and Future Work}

We revealed two pathological cases in the videos synthesized by the state-of-the-art MoCoGAN model, namely \emph{freezing} and \emph{looping}.
To quantify the temporal diversity, we proposed an interpretable class of metrics.
We showed that the SSIM- and PSNR-based metrics, \tpsnr and \tssim, effectively complement IS and FVD to quantify temporal artifacts.
Next, we traced the artifacts to the limited training length, which inhibits long-term modeling of the video sequences.
As a remedy, we reformulated video generation as an MDP and incorporated it into MoCoGAN.
We showed the efficacy of our MDP model on the challenging UCF-101 dataset both in terms of our temporal metrics, as well as in IS and FVD scores.
Maintaining the recurrent state between the training iterations or imposing a tractable prior on the state suggest promising extensions of this work toward generating long-sequence videos.

{\small
\myparagraph{Acknowledgements.} The authors thank Sergey Tulyakov and Masaki Saito for helpful clarifications.
}

{\small
\balance
\bibliographystyle{ieee_fullname}
\bibliography{egbib}
}

\clearpage

\pagenumbering{roman}
\appendix

\title{Markov Decision Process for Video Generation\\\large -- Supplemental Material --}
\author{Vladyslav Yushchenko$^{1*}$ \qquad Nikita Araslanov$^2$ \qquad Stefan Roth$^2$\\
$^1$iNTENCE automotive electronics GmbH, $^2$TU Darmstadt
}

\maketitle

\nobalance

%
%
\section{Overview}

We elaborate on the evaluation protocol used in our study, as well as provide additional qualitative examples both from our approach and MoCoGAN \citesupp{MoCoGAN}.
To enable reproducibility of our approach, we detail the architecture of our MDP-based video discriminator and the training specifics of our MDP approach.

%
%

\section{A Note on Reproducibility}
\label{sec:reproducibility}

In the main text, we indicated a discrepancy between the Inception Score (IS) we attained on UCF-101~\citesupp{UCF-101} and the IS reported in the original work~\citesupp{TGAN, MoCoGAN}.
Recall that we compute the IS following Saito \etal~\citesupp{TGAN}, who trained the C3D network~\citesupp{C3D} on the Sports-1M dataset \citesupp{Sport1M} and then further finetuned it on UCF-101 \citesupp{UCF-101}.
We calculate the IS by sampling the first 16 frames from 10K videos and determining the mean and variance over 4 trials.
As in previous work~\citesupp{TGAN, MoCoGAN}, we use the first training split of the UCF-101 dataset.\footnote{More details on the splits of the UCF-101 dataset are available at \url{http://crcv.ucf.edu/data/UCF101.php}.}
We use original authors' implementation, for both MoCoGAN\footnote{MoCoGAN repository provided at \url{https://github.com/sergeytulyakov/mocogan}.} and TGAN\footnote{Code repository by \citesupp{TGAN}, provided at \url{https://github.com/pfnet-research/tgan}.} and train the respective models for 100K iterations.
We did not observe further improvements of the IS for longer training schedules.

To facilitate transparency and reproducibility of our experiments, we highlight two contributing factors that we carefully considered in our evaluation: \emph{IS implementation} and \emph{model selection}.

\begin{figure}[t]
\centering
\small
\begin{subfigure}{\linewidth}
\removelatexerror
\begin{algorithm}[H]
  \DontPrintSemicolon
  \SetAlgoLined
  \SetKwData{VideoA}{$X_{64 \times 64}$}
  \SetKwData{VideoB}{$\hat{X}_{128 \times 128}$}
  \SetKwData{VideoBN}{$X_{128 \times 128}$}
  \SetKwData{VideoC}{$X_{112 \times 112}$}
  \SetKwData{Scale}{scale}
  \SetKwData{Norm}{norm}
  \SetKwData{Crop}{crop}
  \SetKwData{Net}{C3D}
  \SetKwData{IS}{$S$}
  \KwData{Generated Video \VideoA $\in \mathbb{R}^{K \times 64 \times 64}$}
  \KwResult{Inception Score \IS $\in \mathbb{R}$}
  Bicubic interpolation \VideoB $\leftarrow \Scale(\VideoA)$\;
  Normalize \VideoBN $\leftarrow \Norm(\VideoB)$\;
  Center crop \VideoC $\leftarrow \Crop(\VideoBN)$\;
  Forward pass \IS $\leftarrow \Net(\VideoC)$\;
\end{algorithm}
\caption{Algorithm-A}
\label{alg:version_a}
\end{subfigure}\\
\vspace{1em}
\begin{subfigure}{\linewidth}
\removelatexerror
\begin{algorithm}[H]
  \DontPrintSemicolon
  \SetAlgoLined
  \SetKwData{VideoA}{$X_{64 \times 64}$}
  \SetKwData{VideoB}{$\hat{X}_{112 \times 112}$}
  \SetKwData{VideoC}{$X_{112 \times 112}$}
  \SetKwData{Scale}{scale}
  \SetKwData{Norm}{norm}
  \SetKwData{Crop}{crop}
  \SetKwData{Net}{C3D}
  \SetKwData{IS}{$S$}
  \KwData{Generated Video \VideoA $\in \mathbb{R}^{K \times 64 \times 64}$}
  \KwResult{Inception Score \IS $\in \mathbb{R}$}
  Bicubic interpolation \VideoB $\leftarrow \Scale(\VideoA)$\;
  Normalize \VideoC $\leftarrow \Norm(\VideoB)$\;
  Forward pass \IS $\leftarrow \Net(\VideoC)$\;
\end{algorithm}
\caption{Algorithm-B}
\label{alg:version_b}
\end{subfigure}
\caption{\textbf{Two options for IS computation.} While the C3D network requires inputs of size $112 \times 112$, the models for video generation compute sequences at resolution $64\times64$. To adapt this output, we can either \subref{alg:version_a} normalize the input at resolution $128 \times 128$ and crop a centered window $112 \times 112$, or \subref{alg:version_b} normalize the video directly at resolution $112 \times 112$. We show that despite a rather subtle difference, these two reasonable approaches lead to a notable deviation in the Inception Score.}
\end{figure}

\begin{figure*}[t]
\begin{floatrow}
\capbtabbox[1.15\linewidth]{%

\small 
\begin{tabularx}{\linewidth}{@{}Xccc@{}}
\toprule
\multirow{2}{*}{Method}
& \multicolumn{3}{c@{}}{\makecell{Inception Score $\uparrow$}}  \\ \cmidrule(lr){2-4}
& \makecell{Reported}
& \makecell{Reproduced \\ \texttt{Algorithm-A}} 
& \makecell{Reproduced \\ \texttt{Algorithm-B}} \\ \midrule

MoCoGAN~\citesupp{MoCoGAN} & 12.42 $\pm$ 0.03 & 11.15 $\pm$ 0.10 & 12.03 $\pm$ 0.07 \\
TGAN-Normal~\citesupp{TGAN} & 9.18 $\pm$ 0.11 & 8.11 $\pm$ 0.07 & 9.90 $\pm$ 0.06 \\
TGAN-SVC~\citesupp{TGAN} & 11.85 $\pm$ 0.07 & 11.91 $\pm$ 0.21 & 14.04 $\pm$ 0.08\\ \bottomrule
MDP (ours) & 11.86 $\pm$ 0.11 & 11.86 $\pm$ 0.11 & 13.00 $\pm$ 0.07 \\ \bottomrule
\end{tabularx}
}{%
\vspace{2.5em}
\caption[]{\textbf{Reproducibility of the Inception Score (IS) on the UCF-101 dataset.} Despite using the original implementation provided by the authors~\citesupp{TGAN,MoCoGAN} we observe a discrepancy between the reproduced IS and the scores reported in the original work. We identify two factors affecting the reproducibility: model selection and IS implementation.
For the IS implementation, we consider \texttt{Algorithm-A} (\cf Figure~\ref{alg:version_a}) and \texttt{Algorithm-B} (\cf Figure~\ref{alg:version_b}).
While \texttt{TGAN-SVC} with \texttt{Algorithm-A} roughly corresponds to the reported values, the opposite is the case for \texttt{MoCoGAN}. Since mixing the results from the two evaluation algorithms changes the ranking, it is essential that the methods are compared based on the same IS implementation and we ensure this in our experiments (\cf \cref{sec:reproducibility} for a more detailed discussion).}
\label{table:Reproduce}
}%
\hspace{-1em}%
\ffigbox[0.9\linewidth]{%
\centering
\includegraphics[width=0.9\linewidth]{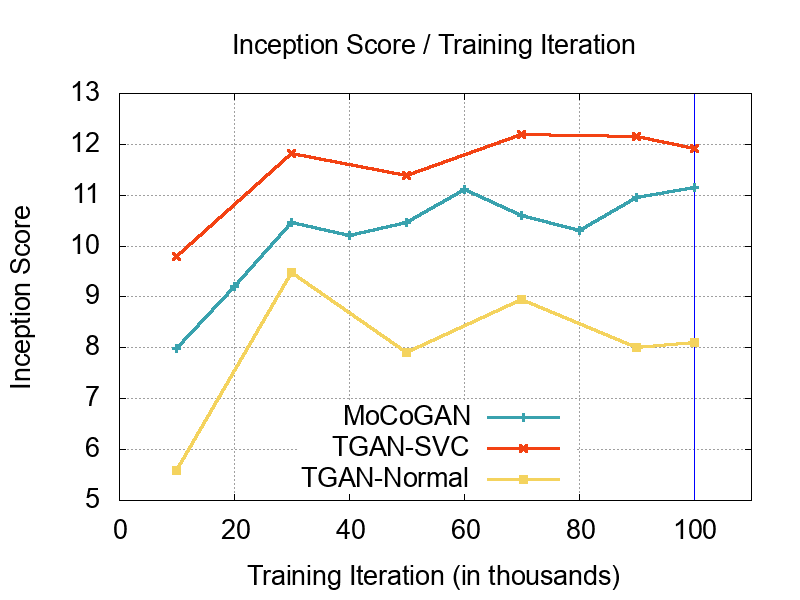}
}{%
	\caption{\textbf{Changes in the Inception Score over the course of training.} Since Inception Score is not the training objective of the GAN models, it is instructive to look at fluctuations in this score over the course of training. We find that longer training does not necessarily improve the IS. Yet, computing the IS is computationally expensive and since there is no train-test split in the conventional sense, an intermediate IS evaluation is equivalent to ``peeking'' into the test performance of the final model. To improve reproducibility, we therefore evaluate the models trained for a fixed number of iterations as indicated by the blue line.}
	\label{fig:plot_is}
}
\end{floatrow}
\end{figure*}

\myparagraph{IS implementation.}
We found the IS to be sensitive to subtle differences in implementation.
Recall that we use the original TGAN~\citesupp{TGAN} evaluation code in our experiments.
\texttt{Algorithm-A} in \cref{alg:version_a} shows the main steps of this evaluation for a single sample of video.
While C3D~\citesupp{C3D} is trained for a resolution of $112 \times 112$, video generators produce sequences of resolution $64 \times 64$.
Additionally, C3D requires the input image sequence to be normalized with the mean and standard deviation used for the network training.
The TGAN evaluation (\cf \cref{alg:version_a}) upsamples the videos to $128 \times 128$, normalizes them, and feeds the center crop of $112 \times 112$  into the C3D network.
However, an alternative evaluation, shown by \texttt{Algorithm-B} in Figure~\ref{alg:version_b}, is to upsample the video directly to $112 \times 112$, applying the normalization at that scale, and feeding the result into C3D without cropping.
This subtle change in the evaluation leads to a tangible difference in the Inception Score.
We summarize the results in Table~\ref{table:Reproduce} and compare these two versions of evaluation to the reported scores in the original works \citesupp{TGAN,MoCoGAN}.

Leaving out cropping for IS computation with \texttt{Algorithm-B} leads to higher values of the IS in comparison to \texttt{Algorithm-A}.
Note that the IS for \texttt{MoCoGAN} produced by \texttt{Algorithm-B} is closer to the reported values: it scores $12.03 \pm 0.07$, which approaches the reported score of $12.42 \pm 0.03$.
However, the opposite is the case for \texttt{TGAN-SVC} as expected, since \texttt{Algorithm-A} is the unaltered version of the evaluation 
provided by the TGAN authors~\citesupp{TGAN} (we attribute the discrepancy for \texttt{TGAN-Normal} to model selection, which we will discuss shortly).
Importantly, regardless of the evaluation protocol our MDP model always outperforms the MoCoGAN baseline and achieves $13.00 \pm 0.07$ with \texttt{Algorithm-B}.

Although the choice of \texttt{Algorithm-B} over \texttt{Algorithm-A} does not change the ranking of the methods, we emphasize that mixing the results produced by the two algorithms does and can essentially invalidate the experimental conclusions.
Therefore, we stick to \texttt{Algorithm-A} for all methods in the experiments that we presented in the main text.

We conclude that the specifics of IS implementation lead to tangibly different results and stress the importance of using the same evaluation strategy for all methods in the experiments.
Moreover, we additionally report the Fr\'echet Video Distance (FVD) as well as our two new metrics, $t$-PSNR and $t$-DSSIM, in the main paper to provide a complementary view to the IS.

\myparagraph{Model selection.}
Recall that the IS for \texttt{TGAN-Normal} using \texttt{Algorithm-A}, $8.11 \pm 0.07$, is still inferior to the reported score of $9.18 \pm 0.11$ (\cf Table~\ref{table:Reproduce}).
To investigate this discrepancy, we observe that the standard training objective \citesupp{GAN} used to train video generation models \citesupp{TGAN,MoCoGAN} serves only as a proxy criterion for the Inception Score.
As a result, lower training loss, or even convergence of training, does not necessarily imply an improvement in the Inception Score.
\Cref{fig:plot_is} illustrates this observation: the Inception Score fluctuates over the course of training, even after a considerable number of training iterations.
Indeed, the IS achieved by \texttt{TGAN-Normal} at around 30K iterations is the highest, $9.48 \pm 0.13$, which is closer to the $9.18 \pm 0.11$ reported by the TGAN authors~\citesupp{TGAN} (in fact better).
Although it is disputable which strategy of selecting the final model for evaluation is more meaningful, we argue against searching for the best IS across training iterations.
One reason is that there is no conventional train-test split for video generation, hence computing the IS for intermediate models amounts to ``peeking'' into the test-time performance of the model.
Selecting the best IS also inhibits reproducibility.
Since IS fluctuations are rather random, there can be no fixed training schedule defined a-priori to reproduce the result.
Additionally, computing the IS is computationally expensive, as it requires thousands of forward passes with a pre-trained classification network (C3D).

We believe that more transparency both at the stage of model selection and IS implementation can improve reproducibility of the Inception Score.
We adhere to this principle and ensure a pre-defined training schedule and exactly same evaluation methodology to enable a fair and reproducible experiments.

\begin{table*}[!t]
\centering
\begin{tabular}{@{}Xcccccccc@{}}
\toprule
\multirow{2}{*}{Model} & \multicolumn{5}{c}{Conv3D} & \multirow{2}{*}{BatchNorm~\citesupp{BatchNorm}} & \multirow{2}{*}{LeakyReLU~\citesupp{maas2013rectifier}} \\ \cmidrule(lr){2-5}
& Filters & Kernel & Stride & Padding & Dilation & & \\ \midrule

\multirow{4}{*}{$D_V$} & 64 & 4,4,4 & 1,2,2 & 0,1,1 & 1,1,1 & \checkmark & \checkmark \\
 & 128 & 4,4,4 & 1,2,2 & 0,1,1 & 1,1,1 & \checkmark & \checkmark \\
 & 256 & 4,4,4 & 1,2,2 & 0,1,1 & 1,1,1 & \checkmark & \checkmark \\
 & 1 & 4,4,4 & 1,2,2 & 0,1,1 & 1,1,1 & &  \\ \midrule

\multirow{4}{*}{TCN} & 64 & 3,4,4 & 1,2,2 & 2,1,1 & 1,1,1 & \checkmark & \checkmark \\
 & 128 & 3,4,4 & 1,2,2 & 4,1,1 & 2,1,1 & \checkmark & \checkmark \\
 & 256 & 3,4,4 & 1,2,2 & 8,1,1 & 4,1,1 & \checkmark & \checkmark \\
 & 1 & 1,4,4 & 1,2,2 & 0,1,1 & 1,1,1 & &  \\ \bottomrule
 
\end{tabular}
\caption[]{\textbf{Original MoCoGAN video discriminator $D_V$ and its TCN architecture adaptation.} The 3D convolution is optionally followed by BatchNorm~\citesupp{BatchNorm} and LeakyReLU~\citesupp{maas2013rectifier} activations, as indicated by the checkmarks. The three parameters for the kernel, stride, padding, and dilation correspond to the temporal and two spatial dimensions (height and width), respectively.}
\label{table_app:TCN}
\end{table*}

%
%
\section{Additional Qualitative Examples}

We make additional qualitative results available at \href{https://sites.google.com/view/mdp-for-video-generation}{https://sites.google.com/view/mdp-for-video-generation}.
The examples provide a visual comparison of MoCoGAN and our MDP-based model on the Tai Chi, Human Actions, and UCF-101 datasets.

%
%
\section{Implementation Details}

\subsection{Architecture}

Recall that our MDP approach is based on an extension of the video discriminator from the original MoCoGAN model to a TCN-like model implementing reward causality.
As Table~\ref{table_app:TCN} details, we only modify the temporal domain and replace the standard convolutions with their dilated variants.
The hyperparameters of the TCN are the number of the dilated layers (blocks) and the convolution kernel size.
Following Bai \etal \citesupp{TCN}, we set both hyperparameters to $3$, since in such configuration the receptive field of the last TCN output covers the entire input sequence of $16$ frames.
Note that the TCN version does not increase the number of parameters of the original $D_V$, but even reduces it due to a smaller kernel size in the temporal domain.
It is only the addition of the Q-value approximation that slightly increases the model capacity of $D_V$.
The architecture of the image discriminator and the generator remain in their original form~\citesupp{MoCoGAN}.

\subsection{Training details}

We follow the training protocol of MoCoGAN \citesupp{MoCoGAN} and use the ADAM optimizer \citesupp{ADAM} for training all networks with a learning rate of $2 \times 10^{-4}$ and moment hyperparameters $\beta_1 = 0.5$, $\beta_2 = 0.999$.
As regularization, we only use weight decay of $10^{-5}$.
We leave the settings for the motion and the content subspaces of MoCoGAN to their default parameters with $d_c = 50$ and the motion dimension set to $d_m = 10$.
We also keep the additive noise $\epsilon \sim \mathcal{N}\left(0\,,\,0.1\right)$ for images fed to the discriminators.

In order to be compatible with the MoCoGAN evaluation (\cf \cref{sec:reproducibility}), all networks are trained for $100K$ iterations with a mini-batch size of $32$, while the seed is kept constant ($0$) to ensure equivalent parameter initialization across the experiments.
For our ablation study on the Tai-Chi dataset, we used a batch size of $64$ and trained for $50K$ iterations.
The training length for the models is set to $K=16$ frames, unless explicitly stated otherwise.
We sample the original data with a fixed \emph{sampling stride} of $2$, as in MoCoGAN training.\footnote{The documentation for training MoCoGAN is provided at \url{https://github.com/sergeytulyakov/mocogan/wiki/Training-MoCoGAN}.}
This means that in order to acquire $16$ frames for training, we extract a $32$ frame sequence with a random starting point from the ground truth dataset and take every second frame.
This procedure increases the amount of motion by reducing the fps of the original video sequence, since \eg on the UCF-101 dataset the originally sampled $16$ frames provide imperceptible changes to scene appearance.

{
\small
\balance
\bibliographystylesupp{ieee_fullname}
\bibliographysupp{egbib_supp}
}

\end{document}